\documentclass{article}

\usepackage[final]{corl_2019} 
\usepackage{amssymb, amsmath, gensymb, bm, commath}
\usepackage{graphicx}
\usepackage{tabularx}
\usepackage[skip=2pt]{caption}
\usepackage{subcaption} 
\usepackage{wrapfig}
\usepackage[dvipsnames]{xcolor}
\graphicspath{ {figures/} }

\newcolumntype{Y}{>{\centering\arraybackslash}X}

\newcommand{\state}{\bm{s}}
\newcommand{\action}{\bm{a}}

\newcommand{\ignorethis}[1]{}

\title{Data Efficient Reinforcement Learning\\ for Legged Robots}

\author{
  Yuxiang Yang, Ken Caluwaerts, Atil Iscen, Tingnan Zhang, Jie Tan, Vikas Sindhwani\\
  Robotics at Google\\
  United States\\
  \texttt{\{yxyang, kencaluwaerts, atil, tingnan, jietan, sindhwani\}@google.com} \\
}

\begin{document}
\maketitle


\begin{abstract}
We present a model-based reinforcement learning framework for robot locomotion that achieves walking based on only 4.5 minutes of data collected on a quadruped robot. 
To accurately model the robot's dynamics over a long horizon, we introduce a loss function that tracks the model's prediction over multiple timesteps.
We adapt model predictive control to account for planning latency, which allows the learned model to be  used for real time control.
Additionally, to ensure safe exploration during model learning, we embed prior knowledge of leg trajectories into the action space. 
The resulting system achieves fast and robust locomotion. Unlike model-free methods, which optimize for a particular task, our planner can use the same learned dynamics for various tasks, simply by changing the reward function.\footnote{A video showing the learning process and results can be found at: \url{https://youtu.be/oB9IXKmdGhc}}
To the best of our knowledge, our approach is more than an order of magnitude more sample efficient than current model-free methods.

\end{abstract}


\section{Introduction}
Robust and agile locomotion of legged robots based on classical control stacks typically requires accurate dynamics models, human expertise, and tedious manual tuning~\cite{kuindersma2016optimization,hutter2017anymal,bledtcheetah,kim2017design}. 
As a potential alternative, model-free reinforcement learning (RL) algorithms optimize the target policy directly and do not assume prior knowledge of environmental dynamics. Recently, they have enabled automation of the design process for locomotion controllers~\cite{hutter2017anymal,tan2018sim, xie2018feedback,actuatornet,stoneicra04}.
Yet, all too often, progress with model-free methods is only demonstrated in simulated environments~\cite{heess2017emergence,peng2017deeploco}, due to the amount of data required to learn meaningful gaits. 
Attempting to take these methods to physical legged robots presents major challenges: Namely, how to mitigate the laborious and time-consuming data collection process~\cite{SAC}, and how to minimize hardware wear and tear during exploration? Additionally, what the robot learns is often a task-specific policy. As a result, adapting to new tasks typically involves finetuning based on new rounds of robot experiments~\cite{MAML}.

In this paper, we propose a model-based learning~\cite{marjaninejad2019autonomous,hafner2018planet,janner2019trust} framework that significantly improves sample efficiency and task generalization compared to model-free methods. 
The key idea is to learn a dynamics model from data and consequently plan for action sequences according to the learned model. 
While model-based learning is commonly considered as a more sample-efficient alternative to model-free methods, its successful application to legged locomotion has been limited~\cite{nagabandi2018neural}. 
Our main challenges are threefold. First, the learned model needs to be sufficiently accurate for long-horizon planning, as an inaccurate model can dramatically degrade the performance of the final controller. This is particularly evident for locomotion due to frequent and abrupt contact events. 
The predicted and real trajectories can quickly diverge after a contact event, even if the single-step model error is small. 
The second challenge is real-time action planning at a high control frequency. To maintain balance, locomotion controllers often run at a frequency of hundreds or even thousands of times per second. Therefore, even a short latency in action planning can significantly affect the performance of controller.
Finally, safe data collection for model learning is nontrivial.
To ensure sufficient exploration, RL algorithms typically drive the actuators using random noise.
However, such random actuation patterns can impose a lot of stress on the actuators and cause mechanical failures, especially during the initial stages of training.

Our proposed algorithm addresses the above challenges. During model learning, we use multi-step loss to prevent accumulation of errors in long-horizon prediction. 
To achieve real-time planning, we parallelize a sampling-based planning algorithm on a GPU. 
Additionally, we plan actions based on a predicted future state using the learned dynamics model to compensate for planning latency.
We develop safe exploration strategies using a trajectory generator~\cite{iscen2018policies}, which ensures that the planned actions are smooth and do not damage the actuators. Combining these three improvements with model-based learning, stable locomotion gaits can be learned efficiently on a real robot. 

The main contribution of our paper is a highly efficient learning framework for legged locomotion. With our framework, a Minitaur robot can successfully learn to walk from scratch after 36 rollouts, which corresponds to 4.5 minutes of data (45,000 control steps) or approximately 10 minutes of robot experimentation time (accounting for the overhead of robot setup). 
To the best of our knowledge, this is at least an order of magnitude more sample efficient than the state-of-the-art on-robot learning method using the same hardware platform~\cite{SAC}. More importantly, we show that the learned model can generalize to new tasks without additional data collection or fine tuning. 

\section{Model-learning and Model-Predictive Control Loop}
We formulate the locomotion problem as a Markov Decision Process (MDP) defined by a state space~$\mathcal{S}$, an action space~$\mathcal{A}$, a state transition distribution~$p(\state_{t+1}|\state_t, \action_t)$, an initial state distribution~$p(\state_0)$ and a reward function~$\mathcal{R}:\mathcal{S}\times\mathcal{A}\to\mathbb{R}$. We apply model-based RL to solve this MDP, which learns a deterministic dynamics model $f_{\theta}$ to approximate $p(\state_{t+1}|\state_t, \action_t)$ by fitting to collected trajectories. The learned model estimates the next state given the current state-action pair, and is used by an action planner to optimize the cumulative reward. 
To account for model inaccuracies, we use a model predictive control (MPC) framework that periodically replans based on the latest robot observation.

\begin{figure}
    \centering
    \includegraphics[trim=0 0 0 3em, clip, width=0.6\textwidth]{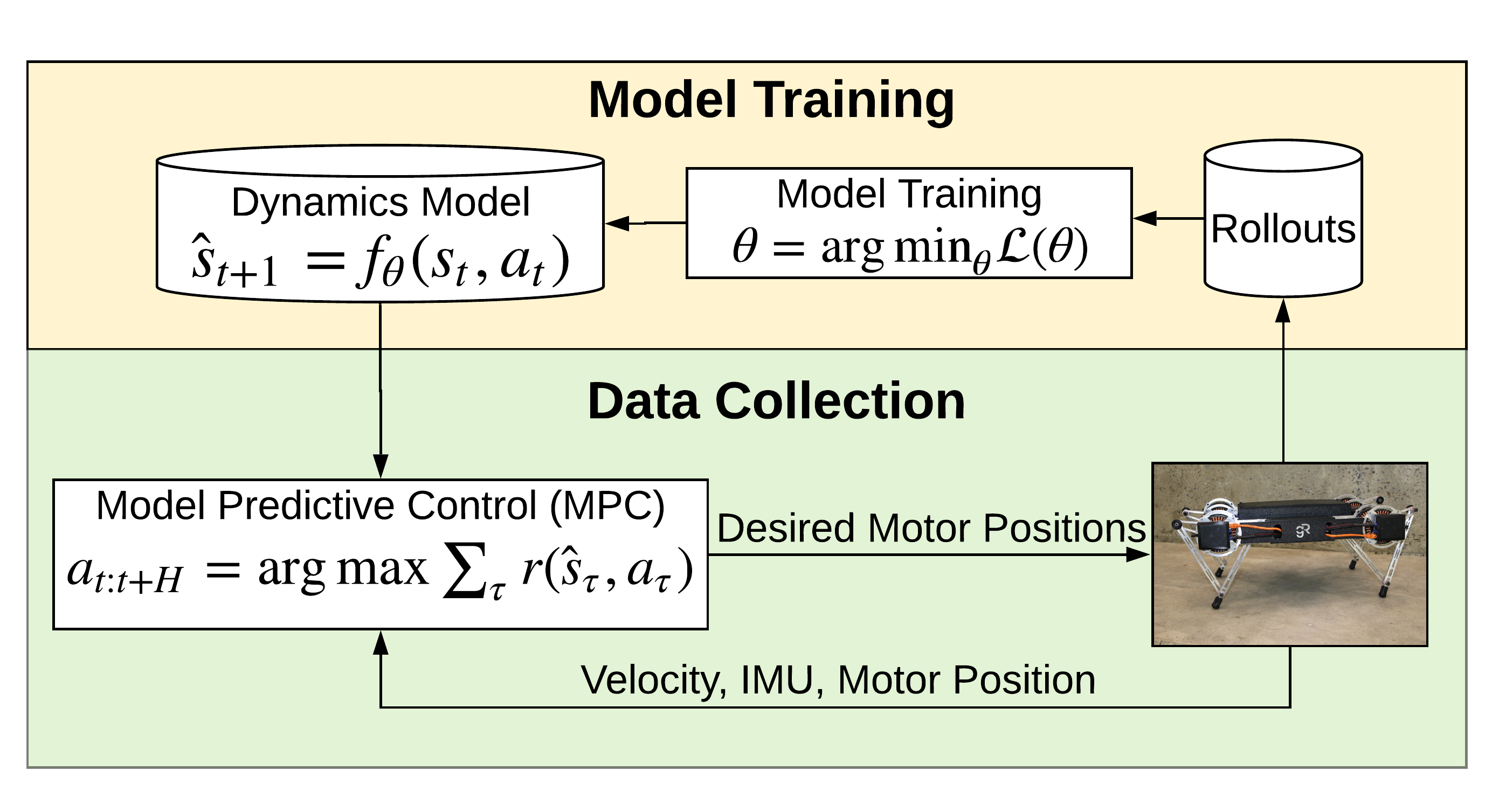}
    \caption{Overview of our learning system on the robot. The system alternates between collecting trajectories and learning a dynamics model.}
    \label{fig:block_diagram}
\vspace{-0.5in}
\end{figure}

Using learned models for action planning raises extra challenges for model accuracy. Although a learned dynamics model can generally remain accurate around trajectories in the training data, its performance for unseen state-actions is not guaranteed. As a result, the planner might exploit such imperfections in the model and optimize for actions that are actually suboptimal on the robot.
To compensate for this distribution mismatch between training and testing data, we keep track of all collected trajectories in a replay buffer and periodically retrain the model using all trajectories \cite{DAGGER}. The updated model is then used to collect more trajectories from the robot, which are added to the replay buffer for future model training (Fig.~\ref{fig:block_diagram}). By interleaving model training and data collection, we improve the model's accuracy on parts of the state space where the planner is more likely to visit, which in turn increases the quality of the plan.

\section{Model-based Learning for Locomotion}
\subsection{Accurate Dynamics Modeling with Multi-step Loss}
We model the difference between consecutive states as a function $f_{\theta}(\state_t, \action_t)=\state_{t+1}-\state_{t}$, where $f_{\theta}$ is a feed-forward neural network with weights $\theta$. Given a set of state transitions $\mathcal{D}=\{(\state_t, \action_t, \state_{t+1})\}$, a standard way to train the model is to directly minimize the prediction error:

\begin{equation}
    \mathcal{L}_{\text{single-step}}(\theta)=\frac{1}{\abs{\mathcal{D}}}    \sum_{(\state_{t}, \action_t, \state_{t+1})\in\mathcal{D}}\norm{(\state_{t+1}-\state_{t})-f_\theta(\state_t, \action_t)}_2^2.
    \label{eq:single-step-loss}
\end{equation}


Although Eq.~\ref{eq:single-step-loss} ensures the model's accuracy for one time step, it does not prevent the accumulation of errors over longer planning horizons.
In previous works, ensembles of models have been exploited to reduce uncertainty and improve the model's long-term accuracy~\cite{handful-of-trials, clavera2018model}. However, prediction using ensembles of models can significantly increase the planning time.
Instead, we use a multi-step loss function~\cite{talvitie2017self, talvitie2014model} to combat the accumulation of model errors. From the current state ${\state}_t$, we use the learned dynamic model $f_\theta$ to predict the next $n$ steps $\{\hat{\state}_{t+1},\hat{\state}_{t+2},...,\hat{\state}_{t+n}\}$ where $\hat{\state}_{t+\tau+1}=\hat{\state}_{t+\tau}+f_\theta(\hat{\state}_{t+\tau})$. Note that the prediction of $\hat{\state}_{t+\tau+1}$ is based on the predicted state of $\hat{\state}_{t+\tau}$, not on the true state $\state_{t+\tau}$, except when $\tau=0$. This is important because this recursively predicted sequence of states allow us to define a multi-step loss function that measures the accumulation of errors over time.
\begin{equation}
    \label{eq:multistep-loss}
    \begin{aligned}
        \mathcal{L}_{\text{multi-step}}(\theta)&=\frac{1}{\abs{\mathcal{D}}}\!\!\sum_{(\state_{t:t+n},\action_{t:t+n-1})\in\mathcal{D}}\!\!\frac{1}{n}\sum_{\tau=1}^{n}\norm{(\state_{t+\tau}-\state_{t+\tau-1})-f_\theta(\hat{\state}_{t+\tau-1}, \action_{t+\tau-1})}_2^2,
    \end{aligned}
\end{equation}
Note that when $n=1$, Eq.~\ref{eq:multistep-loss} reduces to the single-step loss. As $n$ increases, the loss focuses on the accuracy of the model over multiple steps, making the learned model suitable for long-horizon planning. We empirically validate the effect of multi-step loss in Section~\ref{section:multistep-loss-ablation}.

\subsection{Efficient Planning of Smooth Actions}
We use a model predictive control (MPC) framework to plan for optimal actions. Instead of optimizing for the entire episode offline, MPC replans periodically using the most recent robot state, so that the controller is less sensitive to model inaccuracies. Since replanning happens simultaneously with robot execution, the speed of the planning algorithm is critical to the performance of MPC.

With handcrafted models, a number of efficient planning algorithms have been tested for robot locomotion~\cite{neunert2018whole, di2018dynamic, apgar2018fast}. However, they either assume a linear dynamics model, or compute model gradients for linear approximations, which is costly to evaluate for neural networks. Instead, we use the Cross Entropy Method (CEM) to plan for optimal actions~\cite{rubinstein2004cross}. CEM is an efficient, derivative-free optimization method that is easily parallelizable and less prone to local minima. 
It has demonstrated good performance in optimizing neural network functions~\cite{handful-of-trials, kalashnikov2018qt} and can handle non-smooth reward functions. To compute an action plan, CEM samples a population of action sequences at each iteration, fits a normal distribution to the best samples, and samples the next population from the updated distribution in the next iteration.

Sampling each action independently in an $H$-step action sequence is unlikely to generate high quality plans. While good plans often consist of actions that are smooth and periodic, time-independent samples are more likely to produce jerky motions, making it difficult for CEM to select smooth actions. 
Instead, we apply a filter to smooth out the noises added to the mean action. Given a filter coefficient $\gamma\in [0, 1]$, we first generate $H$ time-correlated  samples $\bm{n}_1, \ldots, \bm{n}_H$ from $H$ i.i.d. normally distributed samples $\bm{u}_1, \dots, \bm{u}_H\sim\mathcal{N}(0,1)^{\dim(\mathcal{A})}$:
\begin{align}
    \label{eq:time-correlated-gaussian}
    \bm{n}_1&=\bm{u}_1\\
    \bm{n}_t&=\gamma \bm{n}_{t-1} + \sqrt{1-\gamma^2}  \bm{u}_t.
\end{align}

Given the mean and standard deviation of the action sequences $\bm{\mu}_t,\bm{\sigma}_t,t\in\{1,\dots,H\}$, the sampled actions are computed as $\action_t=\bm{\mu}_t+\bm{\sigma}_t\circ\bm{n}_t$. Note that each sampled action $\action_t$ still follows the desired normal distribution $\mathcal{N}(\bm{\mu}_t, \mbox{diag}(\bm{\sigma}_t))$. However, the exploration noise in consecutive samples is now time-correlated, favoring smooth actions that are less likely to damage the actuators.

\subsection{Online Replanning in the Presence of Latency}
\begin{figure}
    \centering
    \begin{subfigure}[b]{0.45\linewidth}   
        \centering
        \includegraphics[width=\linewidth]{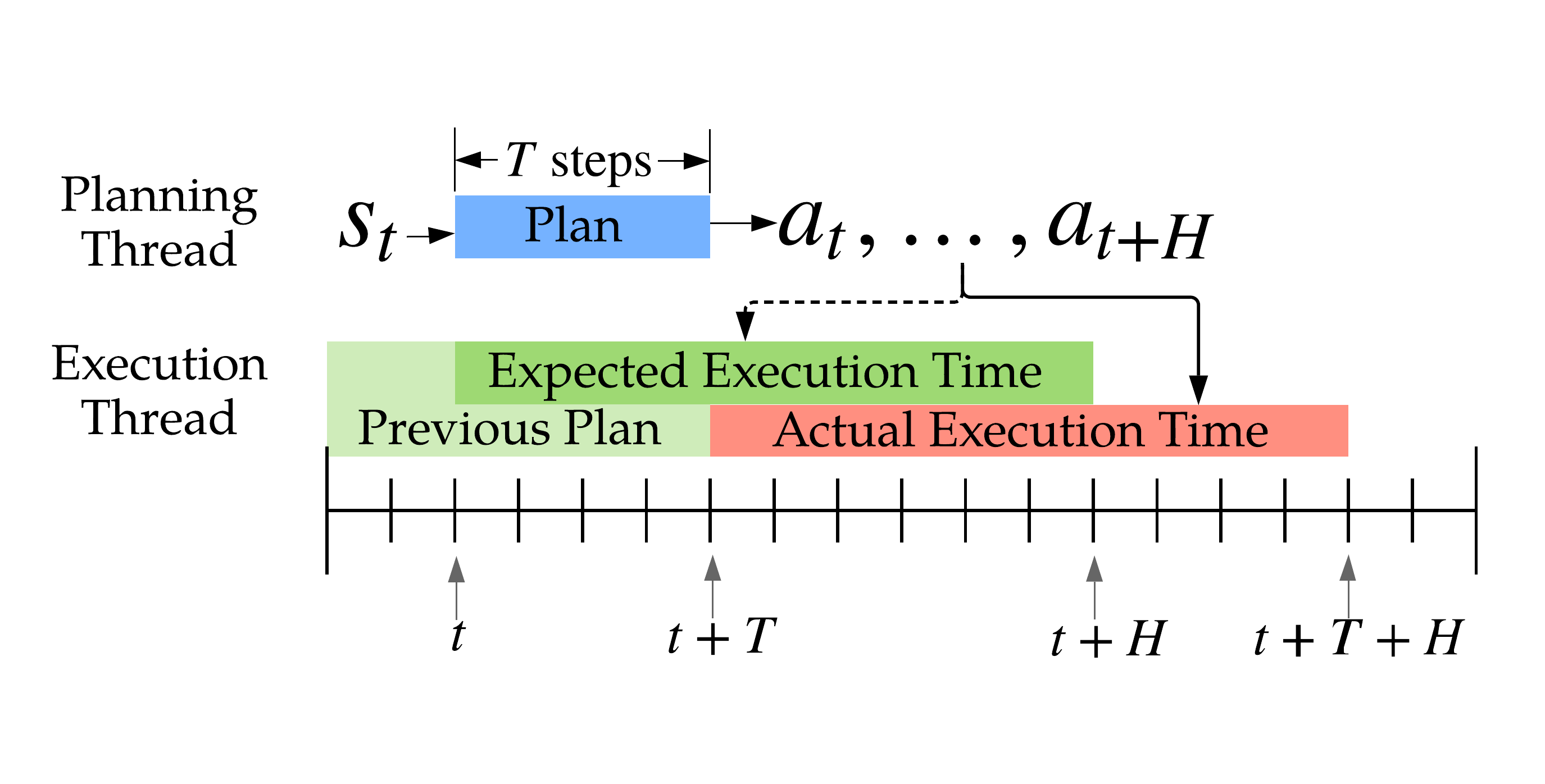}
        \caption{Without asynchronous control.}
        \label{fig:timing_naive}
    \end{subfigure}
    \begin{subfigure}[b]{0.45\linewidth}   
        \centering
        \includegraphics[width=\linewidth]{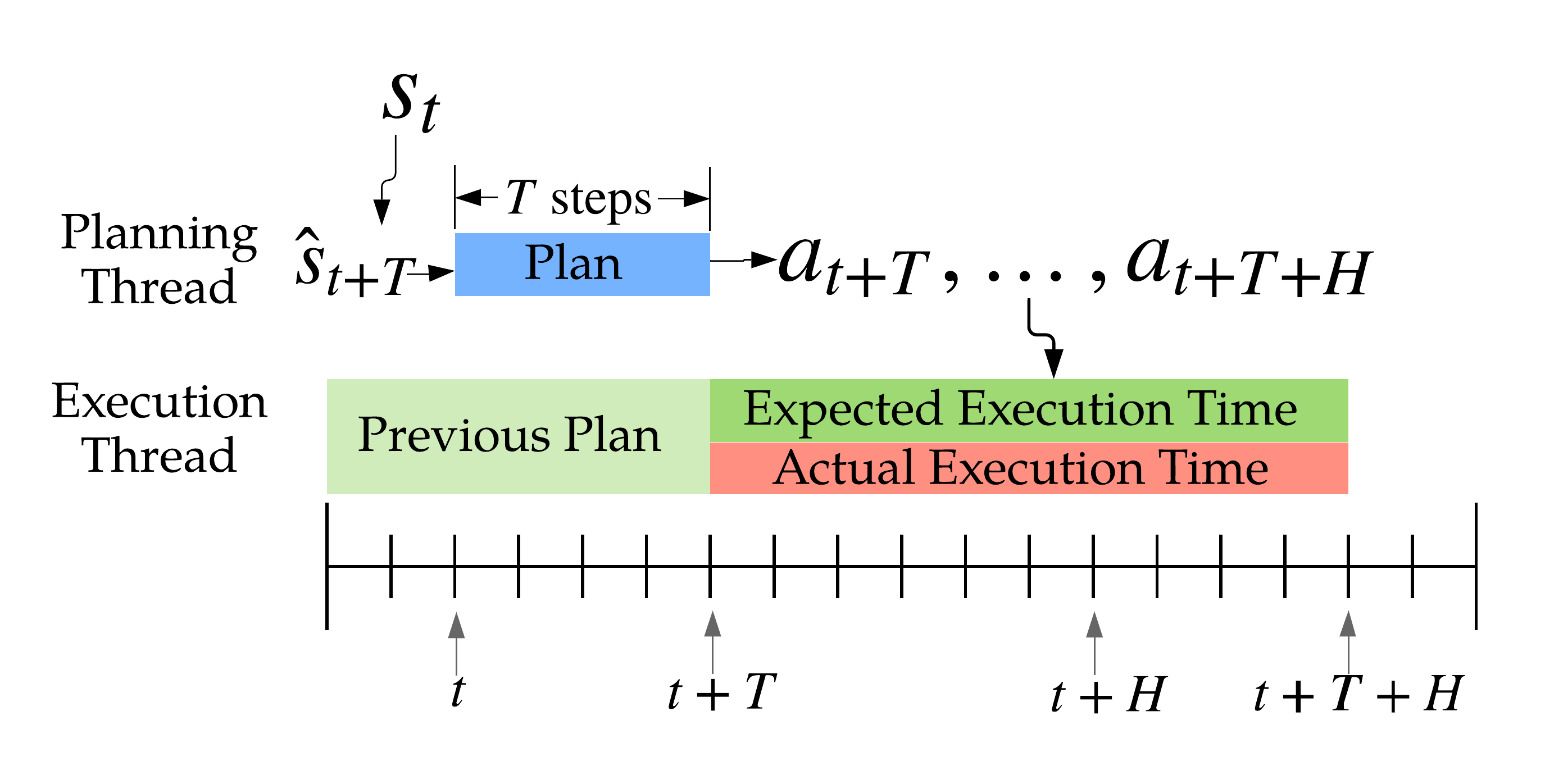}
        \caption{With asynchronous control.}
        \label{fig:timing_async}
    \end{subfigure}
    \label{fig:async_control}
    \caption{Timing diagram of our asynchronous controller. (a) The planner outputs $\action_t, ..., \action_{t+H}$ given the current state $\state_t$. However, due to the planning latency $T$, the action $\action_t$ is actually executed at time $t+T$, on the unplanned state $\state_{t+T}$, which leads to a suboptimal action. (b) Our system predicts the state $\hat{\state}_{t+T}$ when the planning completes, and uses it as the input to the planner. As a result, the planned action $\action_{t+T}$ is executed on the right state $\state_{t+T}$.}
\vspace{-0.15in}
\end{figure}

In classic MPC, replanning happens at every timestep, and only the first action of the planned action sequence is executed. As a result, the planning frequency dictates the control frequency. Since planning usually takes much longer than execution, the reduced control frequency will severely limit the capabilities of the controller. 
We decouple the planning and the control frequencies by parallelizing these two loops. While the planning thread is optimizing for actions in the background, the execution thread simultaneously applies the actions computed in the previous planning step at a higher control frequency. 

Another problem of the long planning time, or the \emph{planning latency}, is that the current state $\state_t$ used by the planner is out-dated when the planning step finishes at $t+T$ (Fig.~\ref{fig:timing_naive}), where $T$ is the computation time needed for planning. In the presence of this latency $T$, we actually apply $\action_t$ at $\state_{t+T}$, which is not optimal. Thus instead of using $\state_t$, we should use $\state_{t+T}$ as the input to the planner. Although $\state_{t+T}$ is a future state that we cannot observe at time $t$, we can predict it using the learned model $f_\theta$ and the actions from the previous planning step. We feed this predicted future state, $\hat{\state}_{t+T}$ to the planner, and the output actions $\action_{t+T},\dots,\action_{t+T+H}$, are then executed when the planning step completes at $t+T$ (Fig.~\ref{fig:timing_async}). This technique, which we call \emph{asynchronous control}, precisely aligns the state that the planner uses and the state when the planned actions will be executed. This provides the planner with a more accurate estimation of the state during execution, and significantly increases the plan quality in the presence of latency. 

\section{Safe Exploration with Trajectory Generators}
\label{section:PMTG}
\begin{figure}[t]
    \centering
    \begin{subfigure}[b]{.2\textwidth}
      \centering
      \includegraphics[trim=0 0 0 0, clip, width=\linewidth]{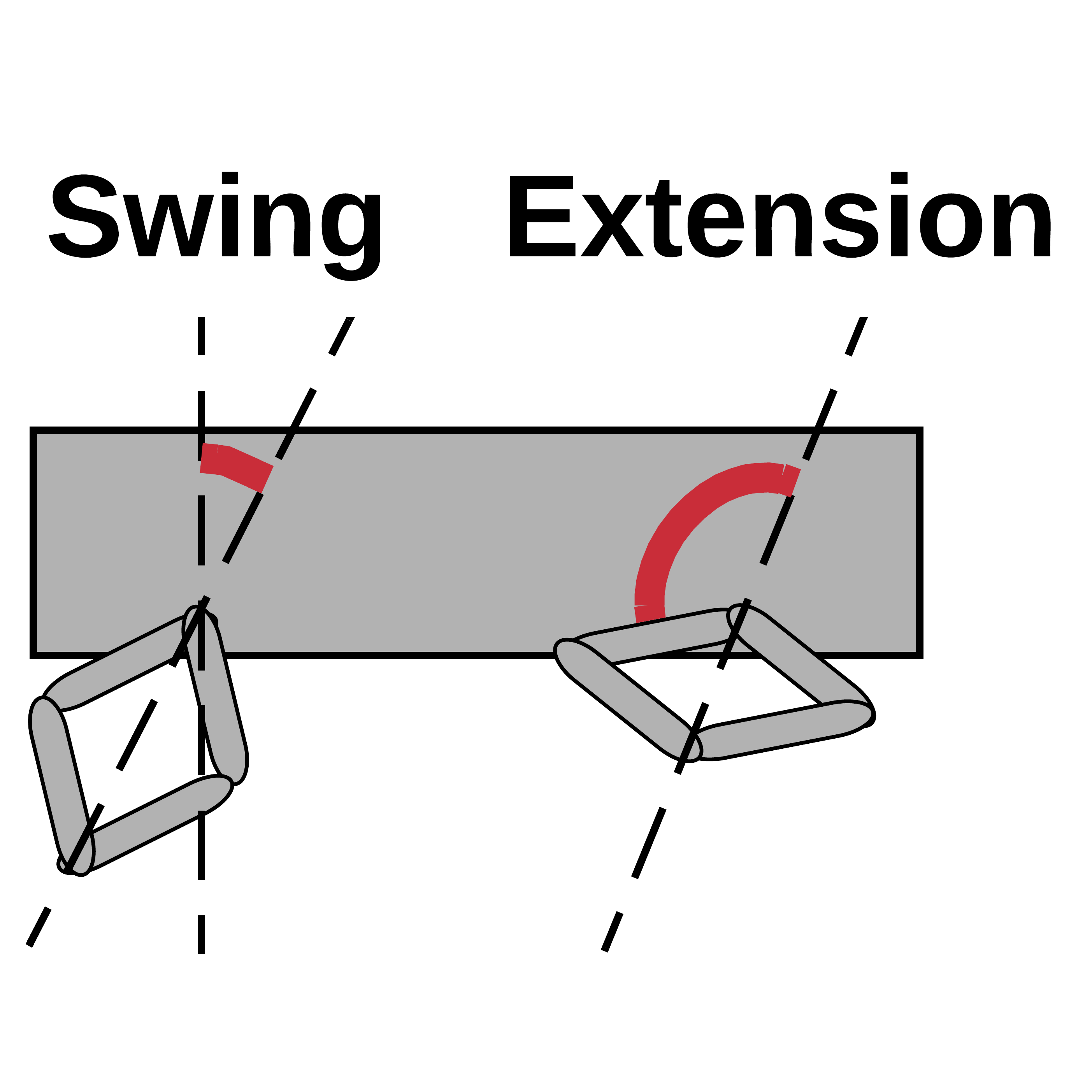}
      \caption{Action space.}
      \label{fig:swing_extension}
    \end{subfigure}
    \begin{subfigure}[b]{.33\textwidth}
      \centering
      \includegraphics[width=\linewidth]{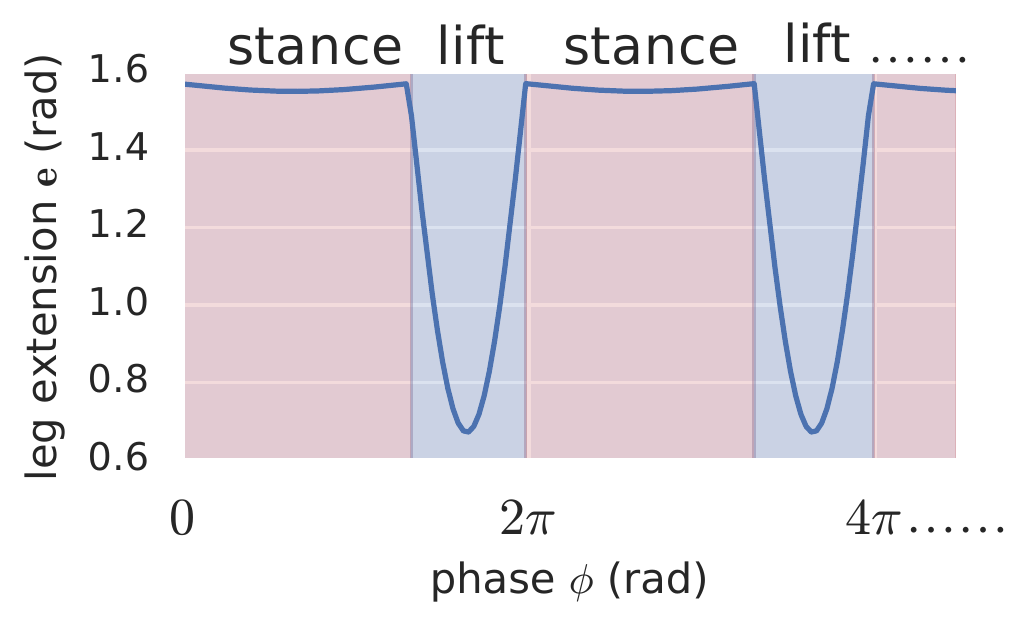}
      \caption{Leg extension of open-loop TG.}
      \label{fig:tg_openloop}
    \end{subfigure}
    \begin{subfigure}[b]{.44\textwidth}
        \centering
        \includegraphics[width=1.0\linewidth]{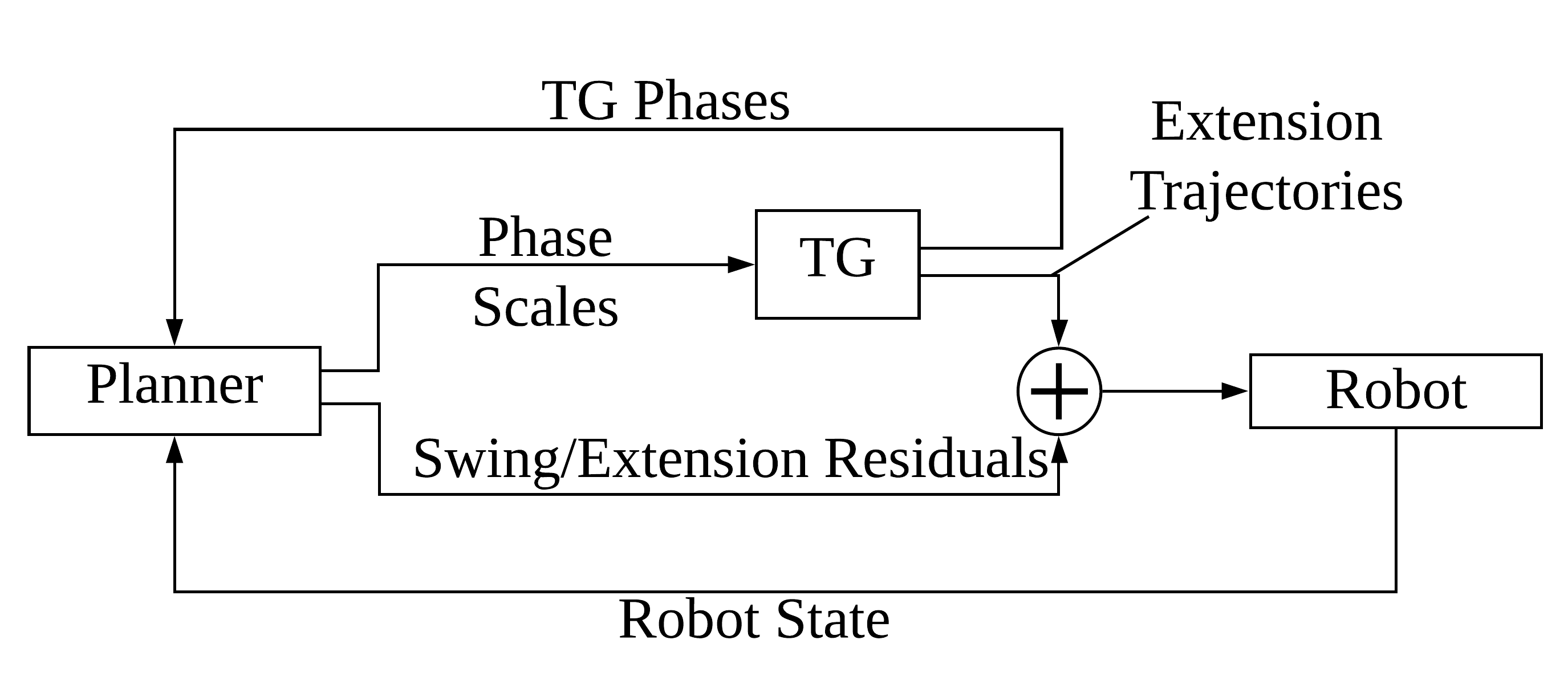}
        \caption{Interaction between planner and TG.}
        \label{fig:tgs_diagram}
    \end{subfigure}
    \caption{Illustration of TGs and their interaction with the planner.}
    \vspace{-1em}
\end{figure}

Whereas formulating the action space using desired motor angles is easier to learn \cite{peng2017learning}, controlling the motors in position control mode can result in abrupt changes in desired motor angles, which may cause large torque output that could potentially damage the robot and its surroundings. 
Instead, we use trajectory generators (TGs) to encourage smooth trajectories.
Similar to~\cite{iscen2018policies}, TGs output periodic trajectories in the extension space of each leg (Fig.~\ref{fig:swing_extension}), and can be modulated by the planner for more complex behaviors.

We use four independent trajectory generators to control all four legs of the robot. Each TG maintains an internal phase $\phi\in [0, 2\pi)$ and controls the leg extension $e$ following a periodic function:
\begin{align}
    e=c_e+a' \cdot \sin(\phi'),\,\text{where }
    a', \phi'=\begin{cases}
        a_{\text{stance}},\frac{\phi}{\phi_{\text{stance}}}\pi &\phi < \phi_{\text{stance}}\\
        a_{\text{lift}}, \left(1 + \frac{\phi-\phi_{\text{stance}}}{2\pi - \phi_{\text{stance}}}\right)\pi&\phi \geq \phi_{\text{stance}}
    \end{cases}
    \label{eq:open-loop-tg}
\end{align}
Here $c_e, a_{\text{stance}}, a_{\text{lift}}, \phi_{\text{stance}}$ are parameters for the TG. As the phase $\phi$ evolves, the TG alternates between the stance mode ($\phi < \phi_{\text{stance}}$) and lift mode ($\phi \geq \phi_{\text{stance}}$). We choose a different amplitude $a_{\text{stance}}, a_{\text{lift}}$ for each mode of the TG, and rescale the original phase to $\phi'$ so that the resulting leg extension is a continuous function. Note that the TGs do not control the leg swing angles. As a result, our planner starts with an open-loop TG that generates an in-place stepping gait~(Fig.~\ref{fig:tg_openloop}). 

We augment the state and action space of the environment so that the planner can interact with TGs~(Fig.~\ref{fig:tgs_diagram}). Our new action space is 12 dimensional. The first 8 dimensions correspond to the swing and extension residual of each leg, which is added to the TG outputs  before the command is sent to the robot. The residuals allow the planner to complement the TG outputs for more complex behaviors. The remaining 4 dimensions specify the phase scales $\omega_{1\dots4}(t)$ for each TG at time $t$, so that the phase of each TG can be propagated independently $\phi_i(t+1)=\phi_i(t) + \omega_i(t)\Delta t$. This gives the controller additional freedom to synchronize arbitrary pairs of legs and coordinate for varied gait patterns. Finally, we augment the state space with the phase of each TG to make the state of TGs fully observable.


\section{Experiments}
\subsection{Experimental Setup}
We use the Minitaur robot from Ghost Robotics~\cite{ghostrobotics} as the hardware platform for our experiments. 
We run our controller with a timestep of 6ms. Similar to \cite{tan2018sim}, the controller outputs desired swing and extension of each leg, which is converted to desired motor positions and tracked by a Proportional Derivative (PD) controller.

We include base linear velocity, IMU readings (roll, pitch, and yaw), and motor positions in the state space of the robot, where the readings come from motion capture (PhaseSpace Inc.~Impulse X2E) and on-board sensors. The state space is 18-dimensional (TG state and sensors). Similar to~\cite{SAC}, we concatenate a history of the past four observations as the input to our dynamics model to account for hardware latency and partial observability of the system. The dynamics are modeled as a feed-forward neural network with one hidden layer of 256 units and tanh activation. We choose $n=20$ as the number of steps to propagate the model and compute the loss.

For MPC, we run CEM for 5 iterations with 400 samples per iteration and a planning horizon of 75 control steps (450 ms). We implement our algorithm in JAX~\cite{jax2018github} for compiled execution and run the algorithm on a Nvidia GTX 1080 GPU. With software and hardware acceleration, our CEM implementation executes in less than 60ms. We replan every 72ms to handle model inaccuracies.

In all experiments where we collect data to train the model, the robot's task is to walk forward following a desired speed profile over an episode of 7.5 seconds. The desired speed starts at zero and increases linearly to a top speed of 0.66 m/s within the first 3 seconds, and remains constant for the rest of the episode. The reward fuction is $r=- \abs{v-\tilde{v}}-0.001 \abs{y}-0.01 (r^2+p^2)$, where $v$ and $\tilde{v}$ are the current and desired walking speed, and $(r, p, y)$ are the roll, pitch, and yaw of the base. The second term encourages walking forward, and the last term stabilizes the base during walking.

\subsection{Learning on Hardware}
\label{section:learning_on_real}
\begin{figure}[t]
    \centering
    \begin{subfigure}[b]{0.45\textwidth}   
        \centering
        \includegraphics[width=0.8\textwidth]{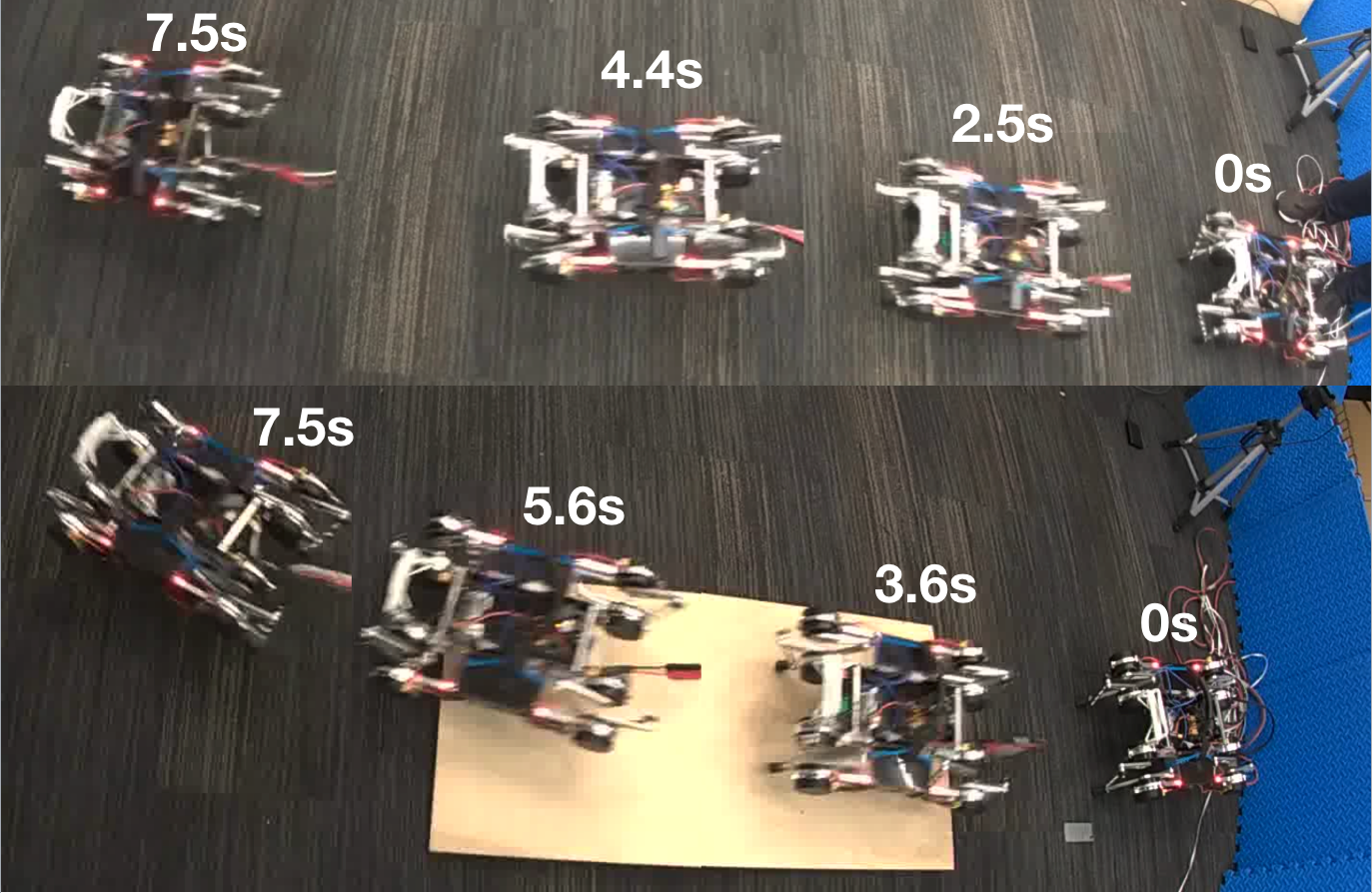}
        \caption{Walk on flat terrain (top) and slope (bottom).}
        \label{fig:real_robot_behavior}
    \end{subfigure}
    \begin{subfigure}[b]{0.45\textwidth}   
        \centering
        \includegraphics[width=\textwidth]{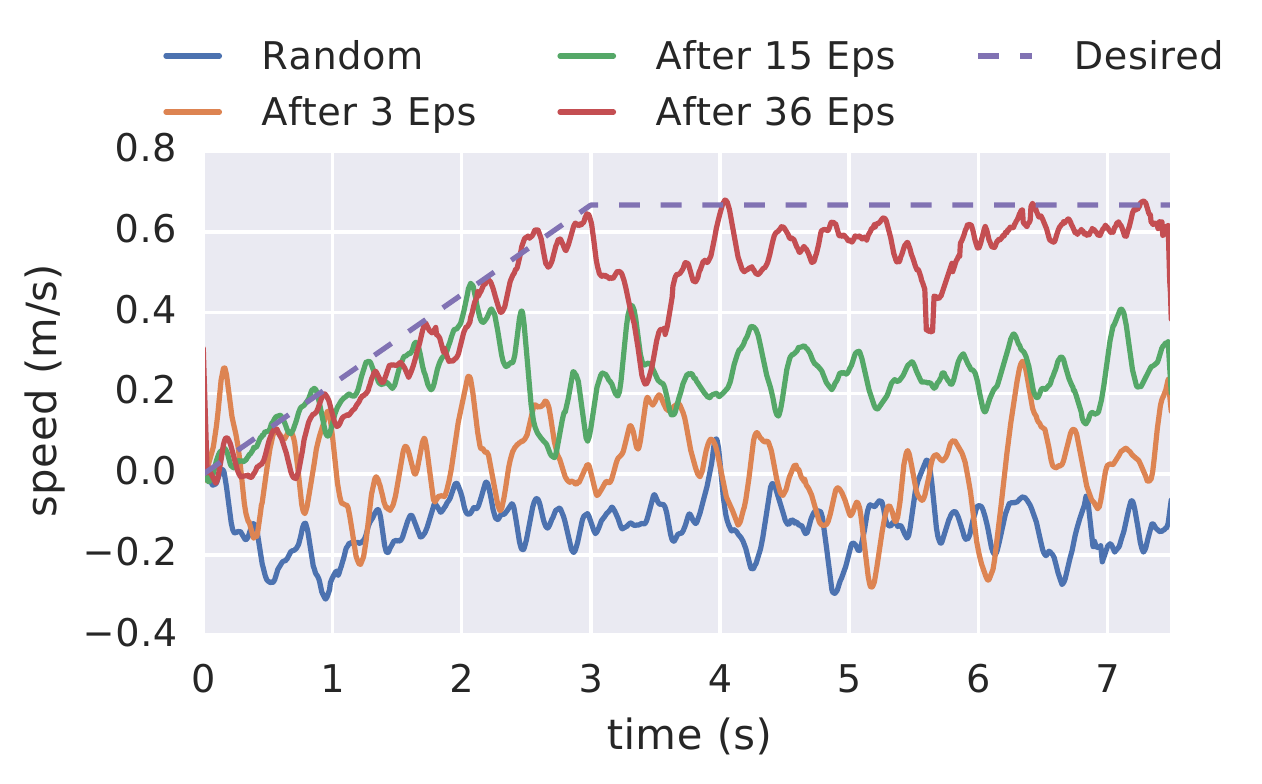}
        \caption{Tracking of desired speed.}
        \label{fig:real_robot_speed_profile}
    \end{subfigure}
    \begin{subfigure}[b]{0.45\textwidth}   
        \centering
        \includegraphics[width=\textwidth]{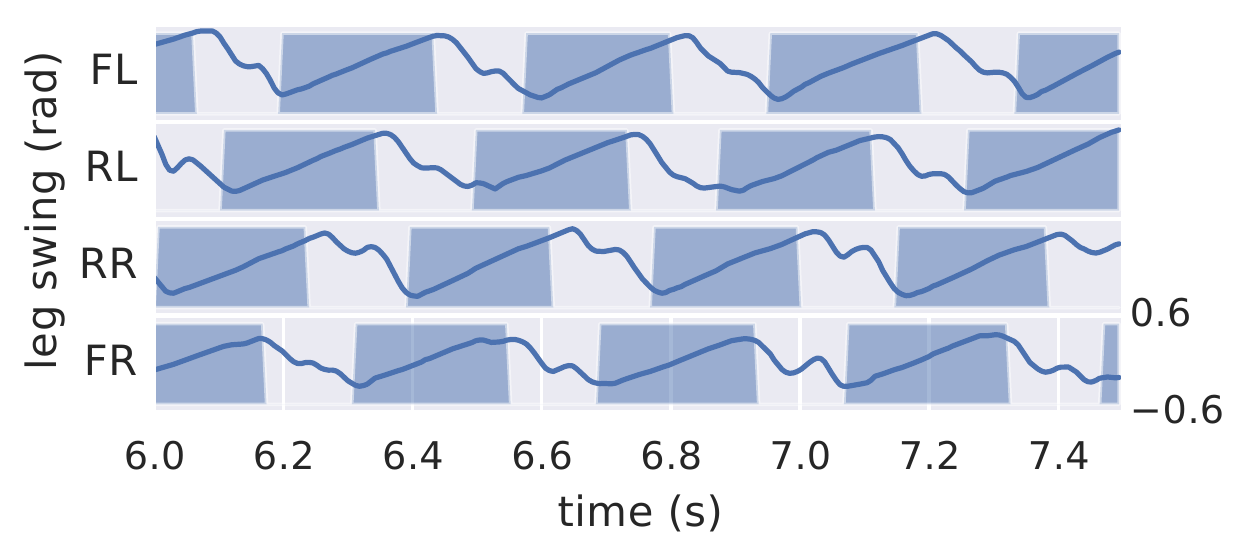}
        \caption{Final gait (dark blue indicates stance phase).}
        \label{fig:real_robot_gait_pattern_final}
    \end{subfigure}
    \begin{subfigure}[b]{0.45\textwidth}   
        \centering
        \includegraphics[width=\textwidth]{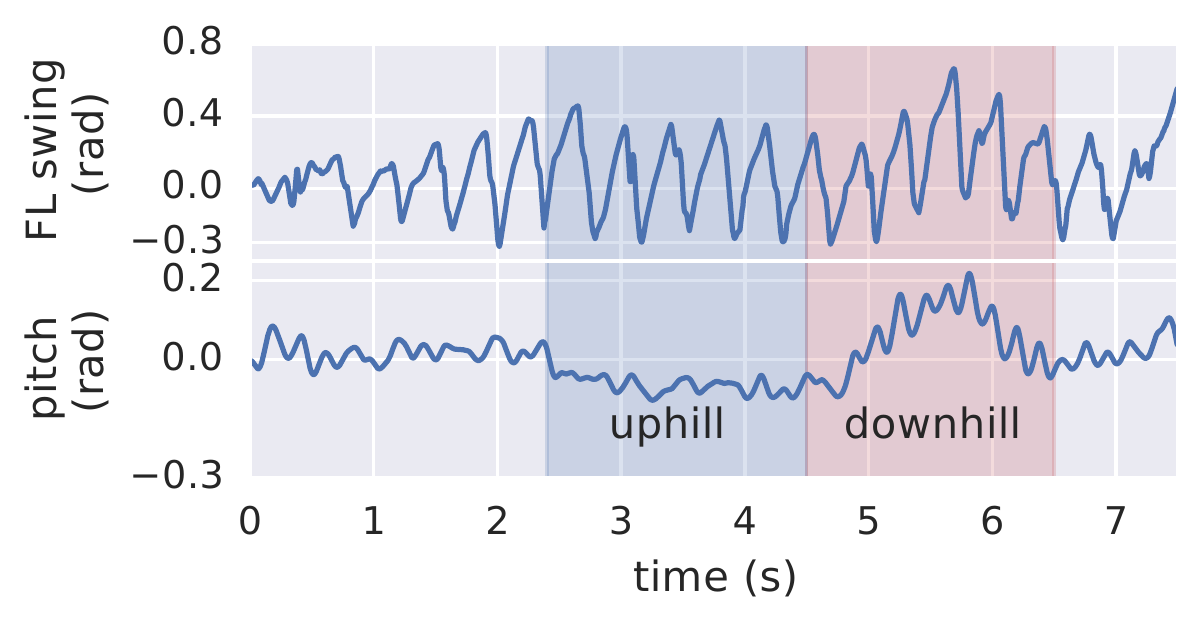}
        \caption{Robot states while walking on a slope.}
        \label{fig:real_robot_perturbation}
    \end{subfigure}
    \caption{Learning on real robot.
    (\ref{fig:real_robot_behavior}) The robot walks on different terrains. (\ref{fig:real_robot_speed_profile}) The robot gradually tracks the desired speed profile. (\ref{fig:real_robot_gait_pattern_final}) Swing angles and gait pattern of all four legs. (\ref{fig:real_robot_perturbation}) Robot trajectory when the robot walks up and down a slope.}
    \vspace{-1em}
\end{figure}
Our method successfully learns a dynamics model based on data from a real robot and optimizes a forward walking gait in only 36 episodes (45,000 control steps), which  corresponds to approximately 10 minutes of robot time, including rollouts, data collection, and experiment resets  (Fig.~\ref{fig:real_robot_behavior} top). 
We update the dynamics model every 3 episodes. 
The robot tracks a desired speed of 0.66 m/s (Fig.~\ref{fig:real_robot_speed_profile}), or 1.6 body lengths per second, which is twice the fastest speed achieved by~\cite{SAC}. The entire learning process, including data collection and offline model training, takes less than one hour to complete. The attached video illustrated the learning process.

It is important to interleave data collection and model training, and update the dynamics model using new data (Fig.~\ref{fig:real_robot_speed_profile}). Initially, when trained only on random trajectories, the model cannot predict the robot dynamics accurately, and MPC only achieves a slow forward velocity. As more data is collected, the model becomes more accurate in the part of the state space which the planner is likely to utilize, leading to better planning performance.

Periodic and distinctive gait patterns develop as the training proceeds (Fig.~\ref{fig:real_robot_gait_pattern_final}). With TGs providing the underlying trajectory, MPC swings the legs forward in the lift phase and backward in the stance phase, leading to a periodic forward-walking behavior. Note that TGs affect leg extensions only, and the leg swing angles are controlled exclusively by MPC. 
Additionally, the ability for MPC to control the phase of each TG allows individual legs to be coordinated. 
In the learned gait, MPC swings the four legs in succession, resulting in a walking pattern.

We also test the robustness of MPC on an unseen terrain. We place a slope in the robot's path (Fig.~\ref{fig:real_robot_behavior} bottom). Although the robot has not trained on the slope, it still can maintain a periodic gait using MPC (Fig.~\ref{fig:real_robot_perturbation}). The robot's pitch angle shows slight perturbations while walking uphill and downhill, but the robot remains upright most of the time.

\subsection{Generalization to Unseen Tasks}
We test the ability of our learned dynamics model to generalize to unseen tasks. We take the dynamics model learned in Section~\ref{section:learning_on_real}, which is trained for walking forward, and perform MPC on new tasks with unseen reward functions. 
For example, to make the robot turn left, we change the reward function to $r=-\abs{v-\tilde{v}}-0.001 \abs{\dot{y}-\widetilde{\dot{y}}}-0.01 (r^2+p^2)$ for a desired turning rate $\widetilde{\dot{y}}$, where $\dot{y}$ is approximated by finite difference.

\begin{figure}[t]
    \centering
    \begin{subfigure}[b]{0.45\textwidth}
        \centering
        \includegraphics[width=\textwidth]{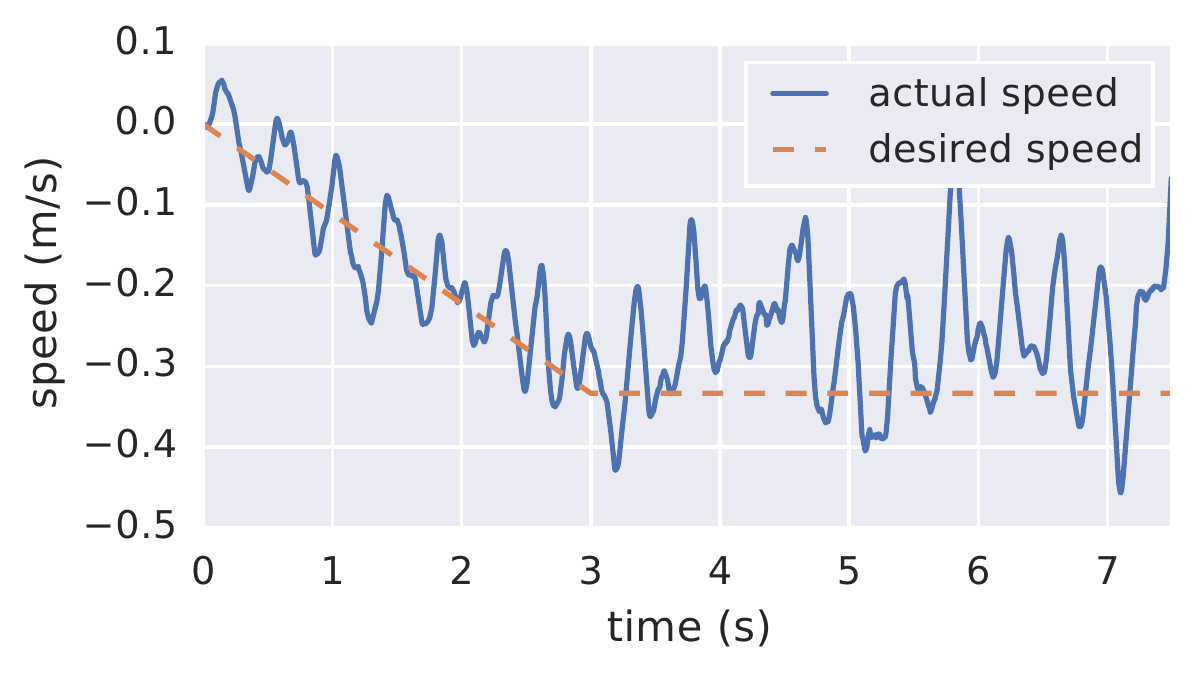}
        \caption{Tracking of a desired backward speed.}
        \label{fig:generalization_backward}
    \end{subfigure}
    \begin{subfigure}[b]{0.45\textwidth}
        \centering
        \includegraphics[width=\textwidth]{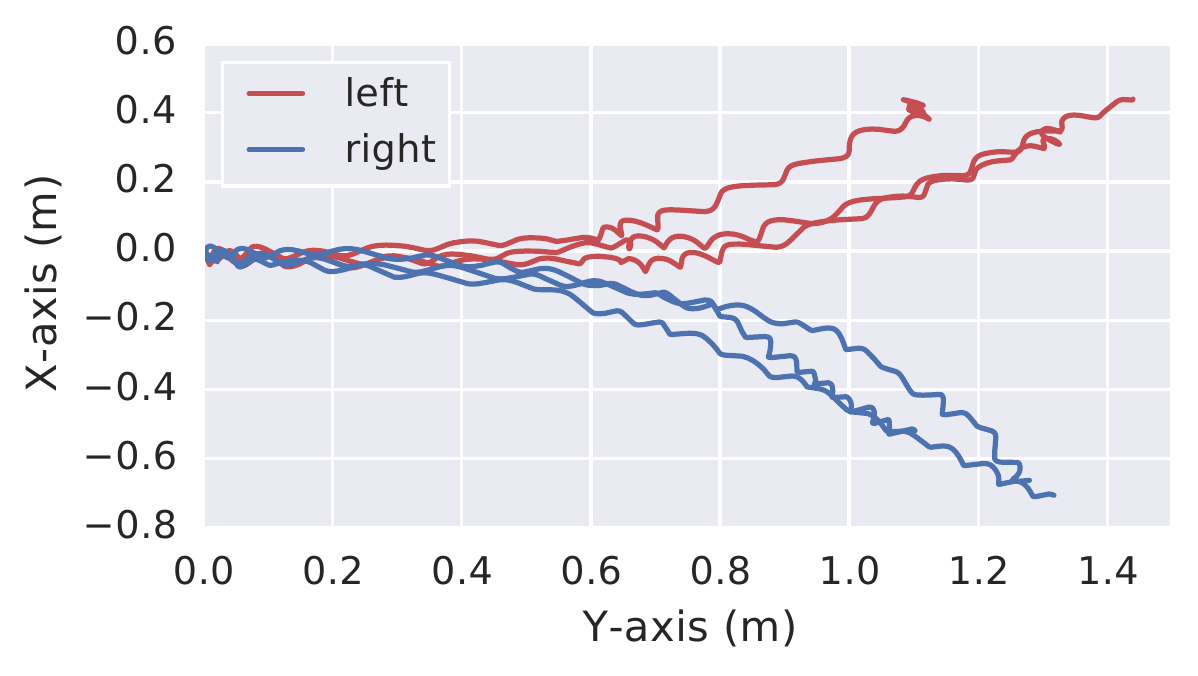}
        \caption{Trajectory of robot turning in x-y plane.}
        \label{fig:generalization_turn}
    \end{subfigure}
    \caption{Generalization of MPC to unseen reward functions using the existing dynamics model. In both cases, the dynamics model is trained only on the task of  forward-walking. In~\ref{fig:generalization_backward}, the new cost function is to track a desired backward speed. In~\ref{fig:generalization_turn}, the new cost function is to keep the same forward speed while turning left or right at a rate of 15 degrees per second.}
\end{figure}

Even though we only train our dynamics model on the task of walking forward, the model is sufficiently accurate to allow MPC to plan for new tasks, including walking backwards and turning (Fig.~\ref{fig:generalization_backward}, \ref{fig:generalization_turn}). 
By learning the dynamics instead of the policy, our algorithm achieves zero-shot generalization to related tasks.

\subsection{Ablation Study}

\begin{figure}[t]
    \centering
    \begin{subfigure}[b]{.45\textwidth}
        \centering
        \includegraphics[width=\linewidth]{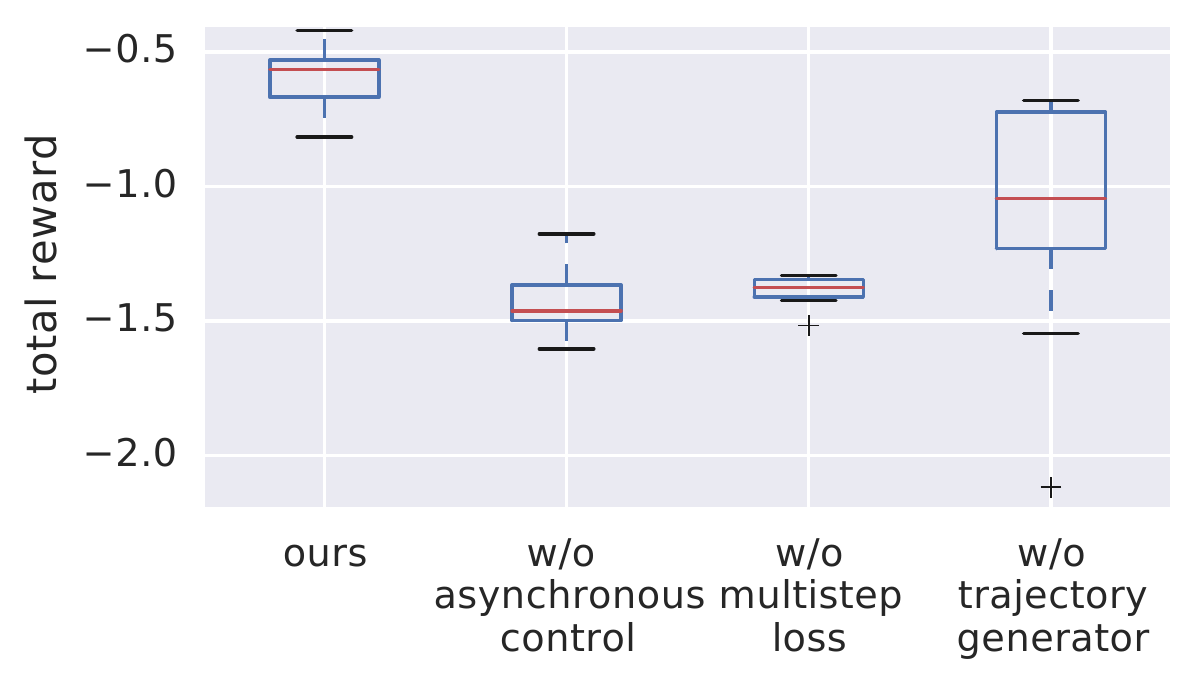}
        
        \caption{\label{fig:ablation_study_learning_curves}Final return using various methods.}
    \end{subfigure}
    \begin{subfigure}[b]{.45\textwidth}
        \centering
        \includegraphics[width=\linewidth]{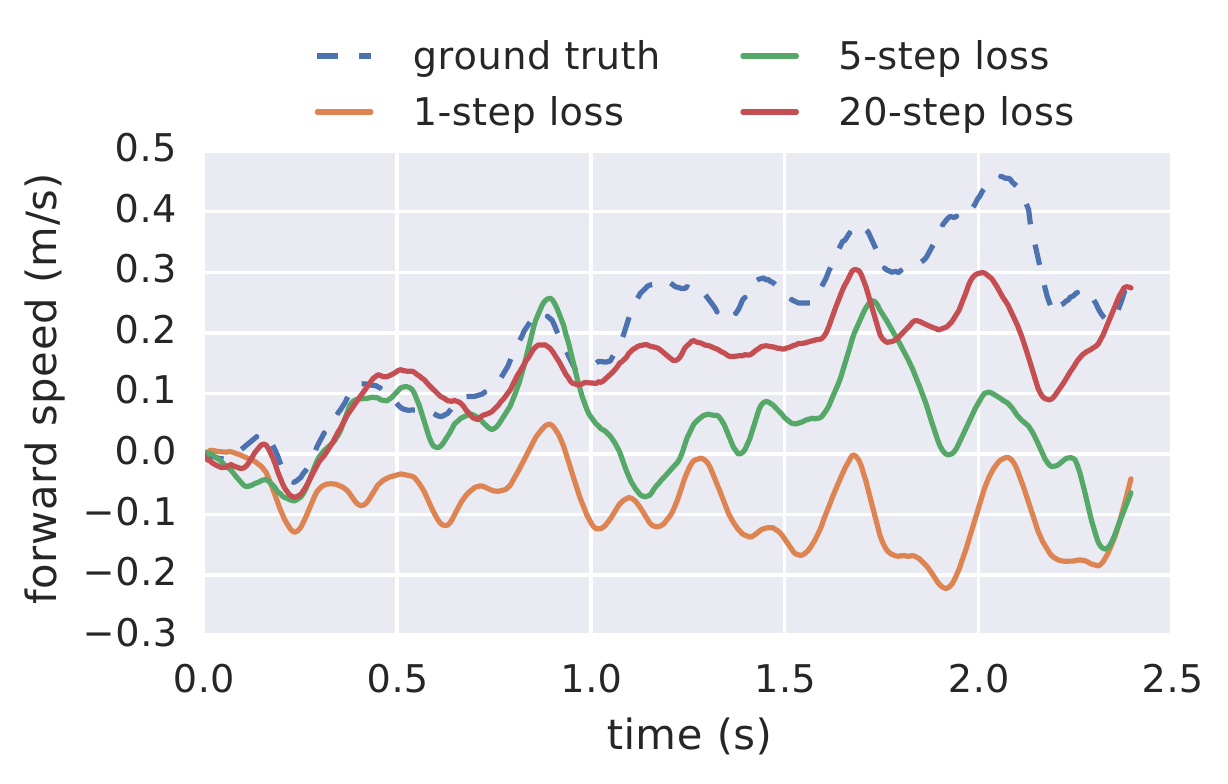}
        \caption{Prediction on unseen trajectory using models trained with different loss functions.}
        \label{fig:ablation_study_single_vs_multi_step}
    \end{subfigure}
    \begin{subfigure}[b]{.45\textwidth}
        \centering
        \includegraphics[width=\linewidth]{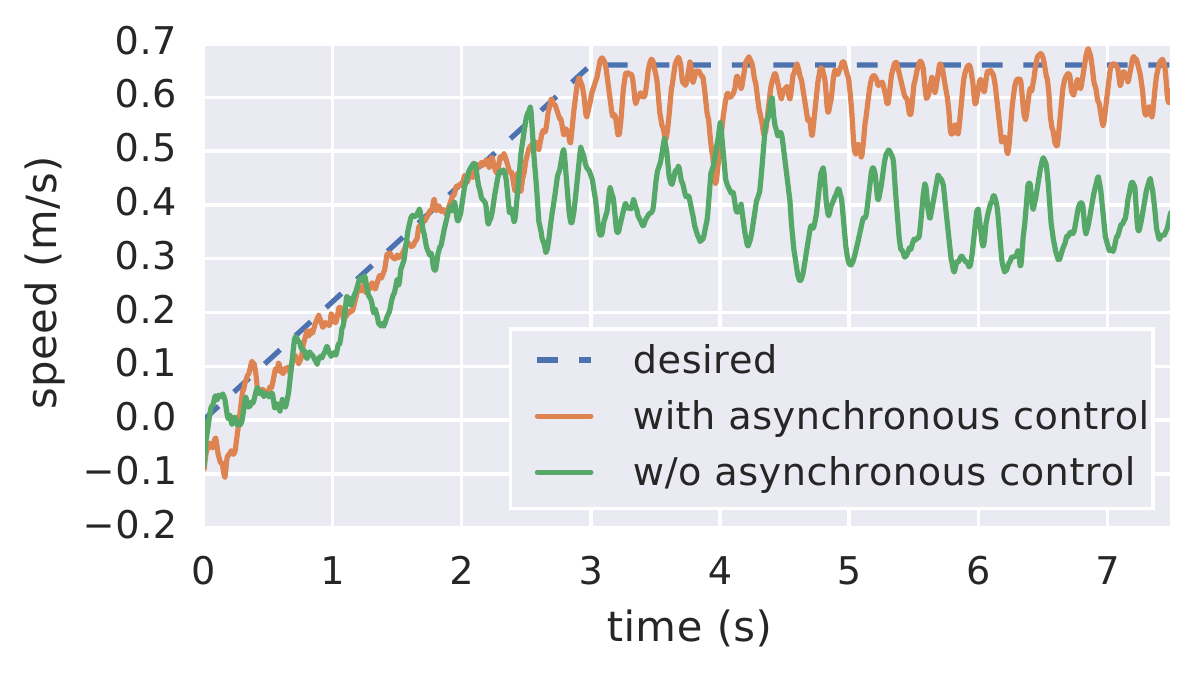}
        \caption{Tracking desired speed with or without asynchronous control.}
        \label{fig:async_control_speed_profile}
    \end{subfigure}
    \begin{subfigure}[b]{.45\textwidth}
        \centering
        \includegraphics[width=\linewidth]{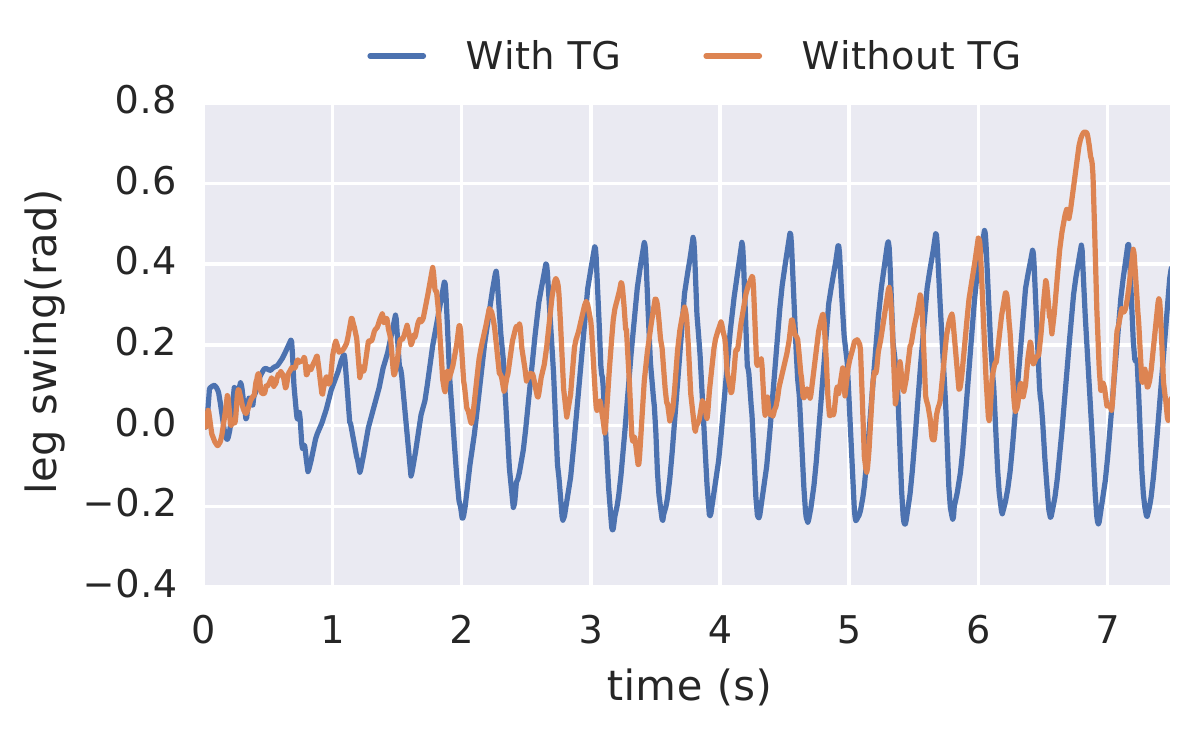}
        \caption{Swing angle of front left leg during MPC, where the model is trained
        with or without TG.}
        \label{fig:ablation_study_reactive_vs_tg}
    \end{subfigure}
    \caption{Ablation study of multistep loss, asynchronous control, and trajectory generators.}
    \vspace{-1em}
\end{figure}
To evaluate the importance of the key components of our algorithm, including multi-step loss, asynchronous control and trajectory generators, we perform an ablation study in a highly accurate, open-source simulation of Minitaur~\cite{tan2018sim}. Simulations help us collect a larger amount of data and reduce the variance of analysis due to algorithmic and environmental stochasticity. 

\subsubsection{Dynamics Modeling with Multi-step Loss}
\label{section:multistep-loss-ablation}
We find that the number of steps ($n$ in Eq.~\ref{eq:multistep-loss}) to compute the model loss $\mathcal{L}_{\text{multi-step}}$ is an important hyperparameter that affects the model accuracy. 
Without multi-step loss, the model cannot accurately track the robot dynamics over a long horizon and the MPC controller does not achieve a high reward (Fig.~\ref{fig:ablation_study_learning_curves}). We further validate this by training models using multiple values of $n$ and testing their performances on an unseen trajectory (Fig.~\ref{fig:ablation_study_single_vs_multi_step}). With more timesteps propagated in computing the model loss, the trained model tracks the ground truth trajectory increasingly better. Note that the plotted state is the velocity of the robot, for which the planner directly optimizes. Inaccurate estimation of the robot velocity is likely to result in suboptimal planning. We choose $n=20$ as a tradeoff between model accuracy and training time.

\subsubsection{Asynchronous CEM Controller}

\begin{wraptable}{R}{0.5\textwidth}
    \centering
    \begin{subtable}{\linewidth}
        \vspace{-1em}

    \caption{Number of CEM iterations.}
    \vspace{-0.5em}
\begin{tabular}{r|rrrr}
\# Iterations & 1     & 3     & \textbf{5}     & 10    \\ \hline
Return        & -1.83 & -0.95 & \textbf{-0.44} & -0.43
\end{tabular}
    ~~\\
    ~~\\
    \end{subtable}
    \begin{subtable}{\linewidth}

    \caption{Smoothing parameter ($\gamma$ in Eq.~\ref{eq:time-correlated-gaussian}).}
    \vspace{-0.5em}
\begin{tabular}{r|rrrr}
Smoothing & 0     & 0.3   & \textbf{0.5}   & 0.9   \\ \hline
Return    & -1.38 & -0.80 & \textbf{-0.44} & -1.65
\end{tabular}
    ~~\\
    ~~\\
    \end{subtable}
    \begin{subtable}{\linewidth}
    \caption{CEM planning horizon (ms).}
        \vspace{-0.5em}
    \begin{tabular}{r|rrrr}
\hspace{1em}Horizon & 150   & 300   & \textbf{450}   & 600   \\ \hline
Return  & -2.44 & -0.40 & \textbf{-0.44} & -0.68
\end{tabular}
    \end{subtable}
    \caption{Ablation study on various parameters of CEM. All rollouts share the same dynamics model. Results show average return over 5 episodes. In bold: selected values.}
    \vspace{-2em}
    \label{tab:cem_ablation}
\end{wraptable}
Planning with asynchronous control is important for fast locomotion (Fig.~\ref{fig:async_control_speed_profile}). 
Without asynchronous control, the MPC controller could only track the desired speed up to approximately 0.4m/s. As the robot moves faster, the robot states can change rapidly even within a few timesteps. Therefore, it is important to perform planning with respect to an accurate  state. This is also illustrated in Fig.~\ref{fig:ablation_study_learning_curves}, where the system struggles to achieve a good final reward without asynchronous control.

We identify additional important hyperparameters for CEM in Table.~\ref{tab:cem_ablation}: CEM requires at least 5 iterations for optimal performance. While smoothing out sampled actions can significantly improve the plan quality, excessive smoothing can make the legs overly compliant for dynamic behaviors. It is important to plan over a sufficiently long horizon to optimize for long-term return. On the other hand, planning for too long makes the planner susceptible to imperfections of the model. 

\subsubsection{Role of Trajectory Generators}
The TGs play an important role in regulating planned actions and ensuring periodicity of leg motion. In Fig.~\ref{fig:ablation_study_reactive_vs_tg}, we compare the final behavior of MPC using models trained with and without TG. While both rollouts achieve similar total reward, planning with TG smooths the motor actions and makes the leg behavior periodic. The learning process is also less stable without TG (Fig.~\ref{fig:ablation_study_learning_curves}). We attempted to learn a model without TG on the real robot. The motors overheated quickly and the jerky motions damaged the motor mounts, forcing us to stop the experiment early.

\subsubsection{Comparison with Model-free Algorithms}
\begin{wrapfigure}{r}{0.5\textwidth}
    \centering
    \includegraphics[trim=0 0 0 0em, clip, width=\linewidth]{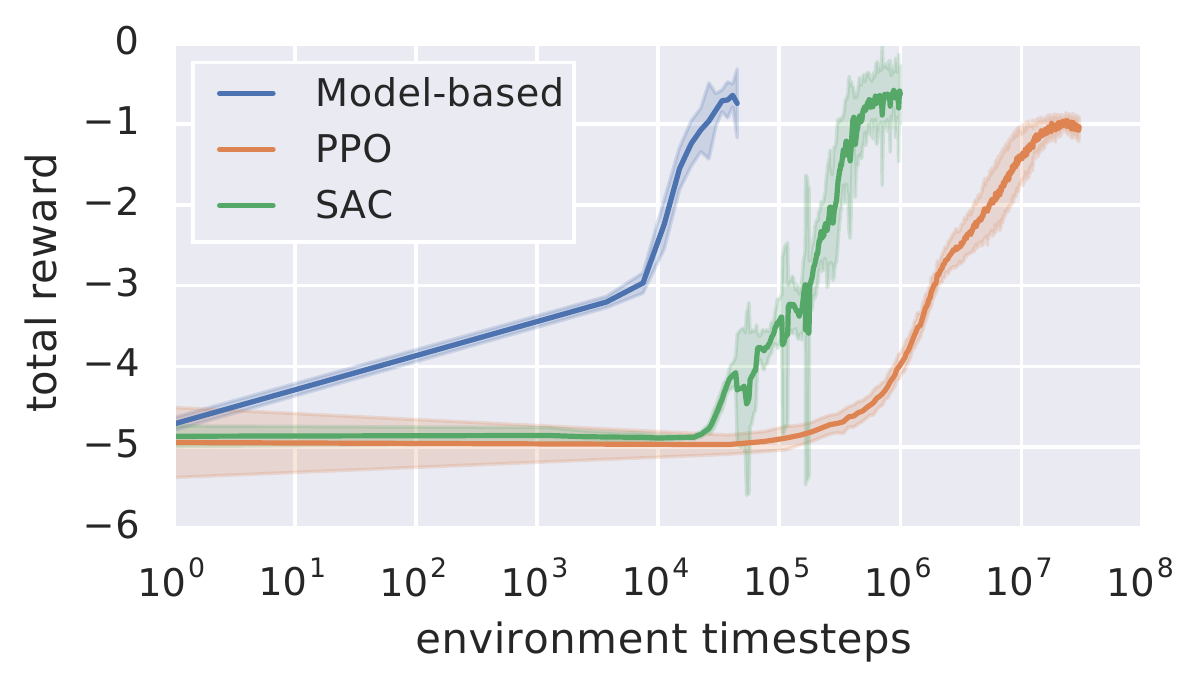}
    \caption{Learning curve of our model-based algorithm compared to model-free ones. Note tha x-axis is on log-scale.}
    \label{fig:model_free_comparison}
    \vspace{-1em}
\end{wrapfigure}
We compare the sample efficiency of our model-based learning method with model-free ones (Fig.~\ref{fig:model_free_comparison}). We obtain the implementations of model-free algorithms from TF-Agents~\cite{TFAgents}. As a state-of-the-art on-policy algorithm, Proximal Policy Optimization (PPO)~\cite{PPO} achieves a similar reward but requires nearly 1000 times more samples, making it difficult to run on the real robot. While the off-policy method, Soft Actor Critic (SAC)~\cite{haarnoja2018soft}, significantly improved sample efficiency and has been demonstrated to learn walking on a Minitaur robot~\cite{SAC}, it still requires an order of magnitude more samples compared to our method, with a less stable learning curve.

\section{Perspectives and Future Work}
The combination of accurate long-horizon dynamics learning with multi-step loss functions, careful handling of real-time requirements by compensating for planning latency, and embedding periodicity priors into MPC walking policies, yields an approach that requires only \mbox{4.5 minutes} of real-world data collection to induce robust and fast gaits on a quadruped robot. Such learning efficiency is more than an order of magnitude superior to model-free methods. The learnt dynamics model can then be reused to induce new locomotion behaviors. 



Yet, many questions remain to be answered in future work: How can rigid-body dynamics be best combined with function approximators for even greater sample efficiency, how should predictive controllers be made aware of model misspecification, and how should predictive uncertainty be best captured and exploited for improved exploration and real-time online adaptation to enable more agile and complex behaviors? Interfacing vision, contact sensing and other perceptual modules with an end-to-end model learning and real-time planning stack is also critical for greater autonomy.


\ignorethis{
Model predictive control has been highly successful for legged locomotion.
However, it can be difficult to define the models for robots with complex dynamics.
In this work we demonstrated sample efficient learning of models that are both accurate and computationally efficient enough to allow for planning based control of legged locomotion.
We made few assumptions about the actuators and sensors and learned approximate forward dynamics using simple feedforward neural network. 

Our main result is that we can learn robust walking at various speeds and turning angles based on only 3 minutes of data collected on the real robot.

[Other future directions by Yuxiang:

More dynamic gaits

Combination with high-level policy for navigation / parkour

Better modeling / planning techniques
]}



\clearpage


\bibliography{corl2019}  
\clearpage

\end{document}